\documentclass[a4paper, 10 pt, conference]{ieeeconf}  

\IEEEoverridecommandlockouts                              

\usepackage{graphics} 
\usepackage{epsfig} 
\usepackage{mathptmx} 
\usepackage{times} 
\usepackage{amsmath} 
\usepackage{amssymb}  
\usepackage{amsfonts}  
\usepackage{subcaption}
\usepackage{bm}
\usepackage{multirow}

\title{\LARGE \bf
Linear Features Observation Model for Autonomous Vehicle Localization}

\author{Oleg Shipitko$^{1}$, Vladislav Kibalov$^{1}$ and Maxim Abramov$^{1}$
\thanks{$^{1}$ Institute for Information Transmission Problems – IITP RAS, Bol’shoy Karetnyy Pereulok 19, Moscow, Russia, 127051; {\tt\small shipitko@iitp.ru}}%
}

\begin{document}

\maketitle
\thispagestyle{empty}

\begin{abstract}
Precise localization is a core ability of an autonomous vehicle. It is a prerequisite for motion planning and execution. The well-established localization approaches such as Kalman and particle filters require a probabilistic observation model allowing to compute a likelihood of measurement given a system state vector, usually vehicle pose, and a map. The higher precision of the localization system may be achieved through the development of a more sophisticated observation model considering various measurement error sources. Meanwhile model needs to be simple to be computable in real-time. This paper proposes an observation model for visually detected linear features. Examples of such features include, but not limited to, road markings and road boundaries. The proposed observation model depicts two core detection error sources: shift error and angular error. It also considers the probability of false-positive detection. The structure of the proposed model allows precomputing and incorporating the measurement error directly into the map represented by a multichannel digital image. Measurement error precomputation and storing the map as an image speeds up observation likelihood computation and in turn localization system. The experimental evaluation on real autonomous vehicle demonstrates that the proposed model allows for precise and reliable localization in a variety of scenarios.
\end{abstract}

\section{INTRODUCTION}

The ability to precisely estimate ego-pose is a core ability of any mobile robot. The motion planning and execution are obstructed by imprecise positioning. The localization problem has been investigated extensively by the research community over the last decades. Visual localization is a localization technique based on data received from visual sensors. Throughout many years it is still one of the core topics in robotics~\cite{lowry2015visual, fuentes2015visual, lu2018survey}. 

The modern visual localization methods are mostly based on detection and matching of surrounding environment features to retrieve motion parameters between images and determining the current camera pose. Among visual features are distinguished global (characterizing the whole image)~\cite{radenovic2016cnn, naseer2018robust} and local ones (characterizing small region of the image)~\cite{qu2016evaluation, saurer2016image}. Global features are known to be less robust against view angle changes and occlusions. On the contrary local features are robust to different image variations. The local features include points~\cite{lowe2004distinctive, krajnik2017fremen}, line segments~\cite{heisterklaus2014image, morago20162d}, contours~\cite{ramalingam2010skyline2gps} and even objects~\cite{fassbender2015landmark}.     

\begin{figure}[thpb]
    \centering
    \includegraphics[width=\linewidth, keepaspectratio]{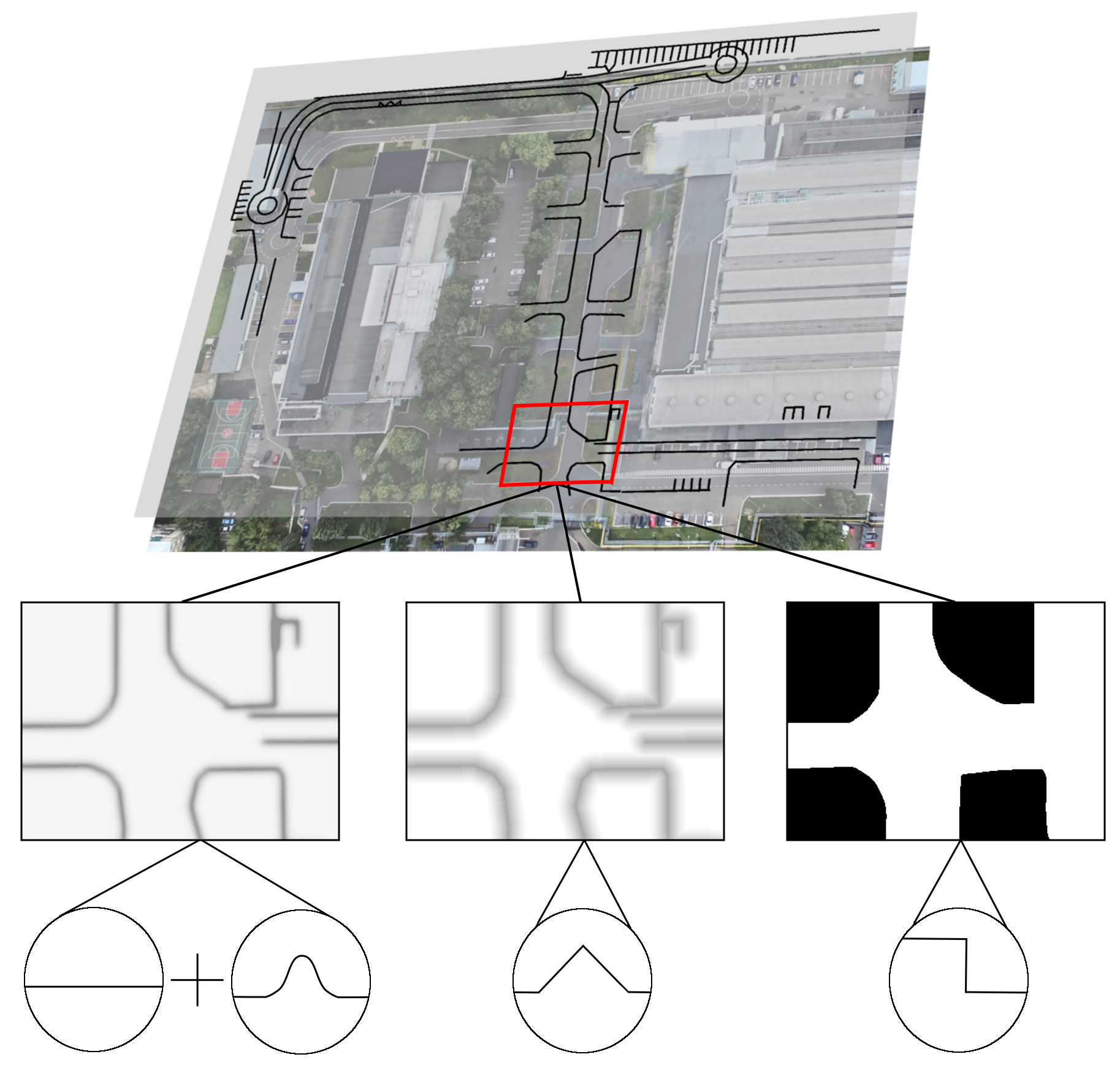}
    \caption{Prior map represented by a multichannel digital image. Each channel encodes a part of a probabilistic observation model.}
    \label{fig:map_model}
\end{figure} 

While point features observation models are widely studied~\cite{piasco2018survey}, there is no rigorous study of the uncertainty sources of linear features visual detection. Although the positioning based on lines detection has a long history~\cite{kahn1990fast}, the need of probabilistic observation model for linear features is reinforced by the growing interest to the self-driving vehicles, since the most obvious positioning data source on the road is road markings which can be represented as a set of linear features. 

By the 2000s the probabilistic robotics became a dominating paradigm in robotics. Probabilistic localization methods are based on the fusion of data received from several sources and account for both motion and measurement uncertainties~\cite{thrun2005probabilistic}. Classical probabilistic data fusion algorithms such as Kalman filter~\cite{wan2000unscented} and particle filter (Monte-Carlo localization)~\cite{dellaert1999monte} require a priory known model of the measurement error distribution. In other words, they use a probabilistic model of sensor measurement (i.e. observation model) allowing to estimate the probability of measurement given the robot pose. It has been long understood that the localization precision is mostly conditioned on the quality of the observation model.

In this work, we propose a probabilistic observation model of visually detected linear features. The proposed model accounts for several major types of measurement uncertainties: shift error, angular detection error, and false-positive detection. We show that the structure of the proposed model allows for efficient precomputation of measurements errors probabilities which can be directly incorporated into the a priori known map. This significantly speeds up the measurement model computation and allows for real-time localization implementation. By experimental evaluation on real vehicles, we demonstrate that the proposed model provides high localization precision in a variety of different localization scenarios. 



\section{RELATED WORK}
Road markings contain essential information used by human drivers for decision making and road situation analysis. It has been long understood that road marking could be used for autonomous vehicle navigation as well. Hillel et al.~\cite{hillel2014recent} provide an extensive overview of lane detection techniques, their limitations, and applications. 

The probabilistic observation model of road markings detector is proposed in~\cite{jo2015precise}. The authors proposed a model accounting for detection shift error. The authors also conducted an experiment evaluating the influence of unpredicted vehicle orientation changes onto the measurement error. They incorporated an error caused by the vehicle orientation changes into the model by increasing the measurement uncertainty according to the magnitude of orientation deviation.

In Bertha (autonomous car made by Daimler AG~\cite{ziegler2014video}) road markings along with point features were used for vehicle localization. The proposed observation model accounts for the decrease of spatial detection precision with the growth of distance between camera and detected road marking.

Monocular visual localization in an urban environment using a particle filter is presented in~\cite{leung2008localization}. In this work, the source of data for localization is the orientation of linear features obtained by analyzing the position of the vanishing point in an image. Detected linear features are compared with the map to estimate the current position of the robot. An exponential heuristic function weighting detected and reference lines misalignment is used to compute measurement likelihood. 

An alternative approach is used in works~\cite{ghallabi2018lidar, chausse2005vehicle}. The authors estimate vehicle position and orientation relative to the detected road markings and compute hypothesis weight as a sum of Gaussian error distributions centered in position and orientation estimates.

There are also works where localization is considered as optimization problem~\cite{lu2017monocular}. Although in this case a probabilistic measurement model is not required, the evaluation results show that this approach can provide necessary localization precision, however, it is less robust in complex environment with partial or full road markings occlusions.

\section{LINEAR FEATURES MEASUREMENT MODEL}

\begin{figure}[thpb]
    \centering
    \begin{subfigure}{0.215\textwidth}
        \includegraphics[width=\linewidth,keepaspectratio]{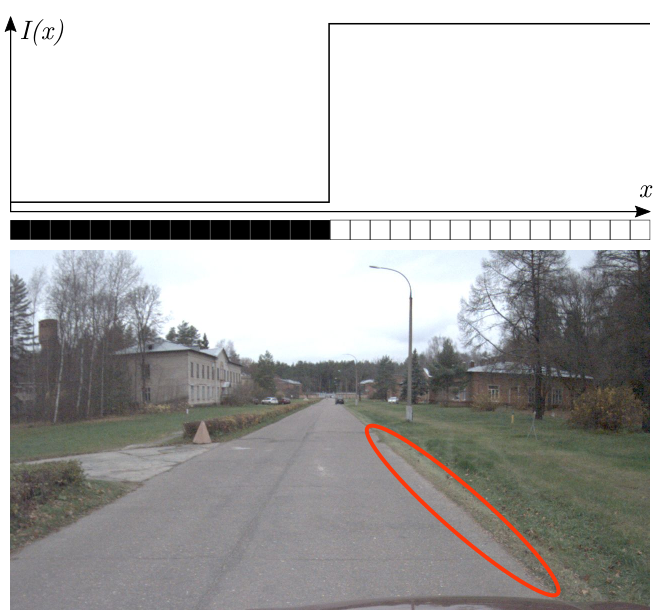}
        \caption{}
    \end{subfigure} 
    \hspace{10px}
    \begin{subfigure}{0.215\textwidth}
        \includegraphics[width=\linewidth,keepaspectratio]{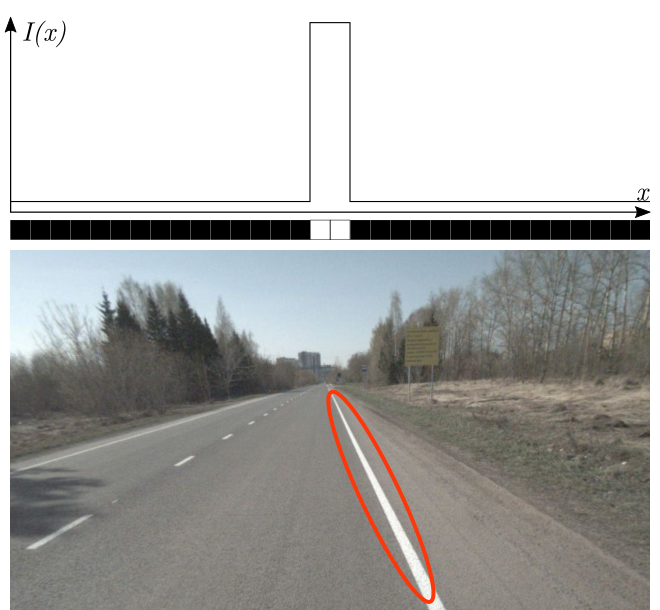}
        \caption{}
    \end{subfigure}
    \caption{Linear feature examples. (a) Image edge: profile (on top), synthetic pixels row (middle), real world example (bottom). (b) Image ridge: profile (on top), synthetic pixels row (middle), real world example (bottom).}
    \label{fig:ridge_edge_example}
\end{figure}

Let us first introduce a concept of a linear feature. Linear feature is any curve detected on image and represented by a set of equidistant points $ l = \{p_i = (x_i, y_i) \mid x_i, y_i \in \mathbb{R}^2, i=1..I, i \in \mathbb{Z}\}$. There are two types of linear features. A step-like image signal change is referred to as edge. A spike-like image change is called a ridge. Examples illustrating the difference between edges and ridges are shown in Fig.~\ref{fig:ridge_edge_example}.

The set of all curves detected on a single image is referred to as a measurement, $ \mathbf{z} = \{l_k,  k = 1...K, k \in \mathbb{Z}\}$. The observation model estimates likelihood $P(\mathbf{z} \mid \mathbf{x}, M)$ of a measurement $\mathbf{z}$ given the vehicle pose $\mathbf{x} = (x, y, \theta)$ and the map $M$. The map in our case is represented by a multichannel digital image containing ground-truth positions of all lines that could be possibly detected (see Fig.~\ref{fig:map_model}). 

\begin{figure*}[thpb]
    \centering
    \includegraphics[width=0.7\textwidth,keepaspectratio]{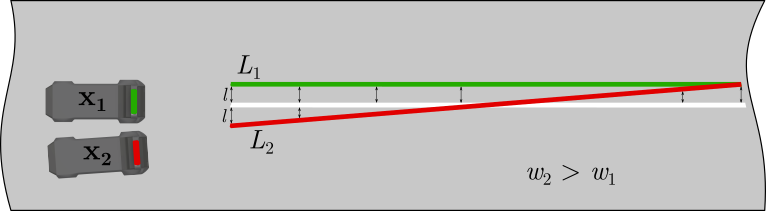}
    \caption{Example of classical model bias. Due to the sampling the incorrect measurement (red) receives higher probability than the correct but slightly shifted one (green).}
    \label{fig:incorrect_meas_example}
\end{figure*}

The observation model accounts for different sources of measurement errors. Classical models for linear features only account for the shift typically exploiting the Gaussian noise model~\cite{jo2015precise}. However, it can be shown that such a model is prone to bias since it does not account for angular misalignment of reference and measured features. Therefore a slightly shifted but correctly oriented line can receive lower likelihood than the line which orientation and shift both estimated incorrectly. An example of this effect is shown in Fig.~\ref{fig:incorrect_meas_example}.

To account for the orientation and overcome such a bias we propose a measurement model that consists of two parts:
\begin{equation}
    P( \mathbf{z} \mid \mathbf{x}, M) = \eta \sum_{k=1}^{K} P_{\mathrm{shift}}(l_k \mid \mathbf{x}, M) \sum_{k=1}^{K} P_{\mathrm{angle}}(l_k \mid \mathbf{x}, M),
    \label{eq:meas_model}
\end{equation}
where $K$ is a number of detected curves, $P_{\mathrm{shift}}$ is a probability density function (PDF) of shift errors measuring how probable the line segment depending on how far it is shifted relative to the expected one and $P_{\mathrm{angle}}$ is a PDF of angular errors showing how probable the detected line is depending on its deviation from the expected orientation, $\eta$ is a normalization coefficient.

\begin{figure}[t]
    \centering
    \begin{subfigure}{0.155\textwidth}
        \includegraphics[width=\linewidth,keepaspectratio]{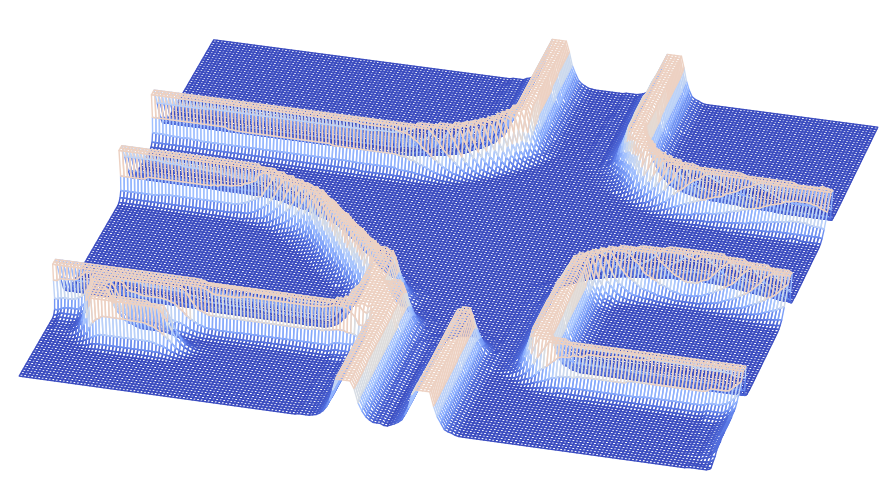}
        \caption{}
    \end{subfigure} 
    \begin{subfigure}{0.155\textwidth}
        \includegraphics[width=\linewidth,keepaspectratio]{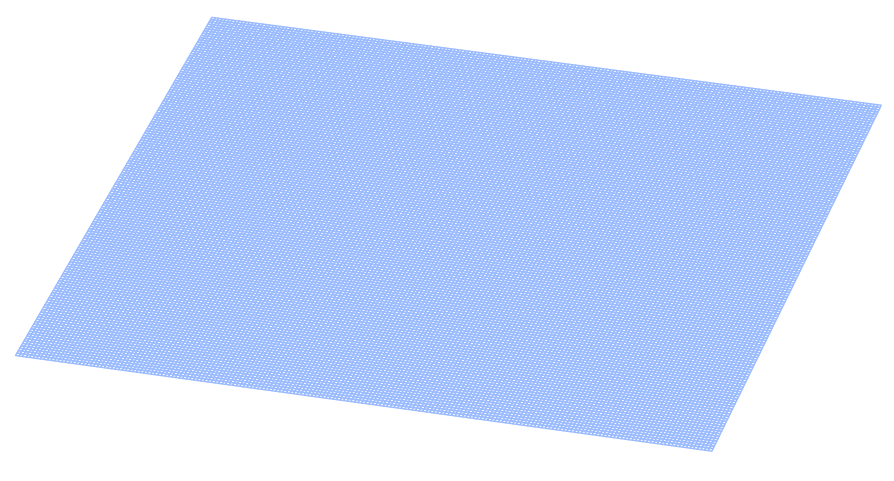}
        \caption{}
    \end{subfigure} 
    \begin{subfigure}{0.155\textwidth}
        \includegraphics[width=\linewidth,keepaspectratio]{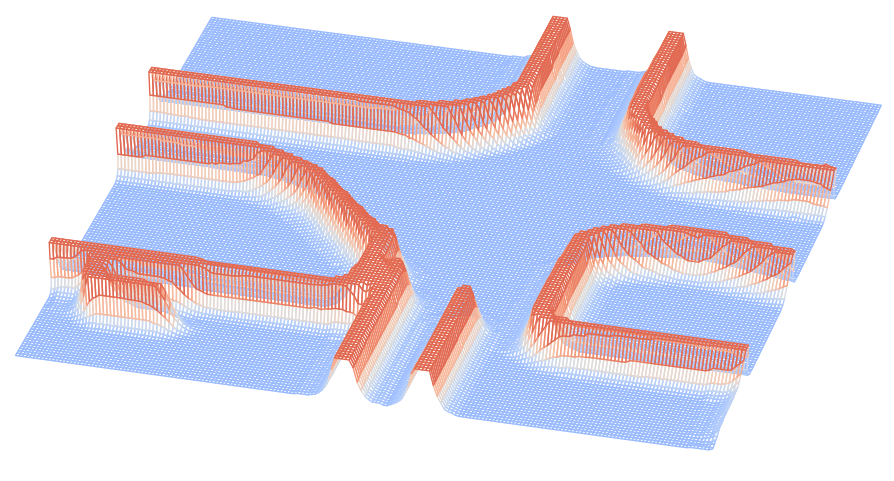}
        \caption{}
    \end{subfigure}\\
    \vspace{10 px}
    \begin{subfigure}{0.48\textwidth}
        \includegraphics[width=\linewidth,keepaspectratio]{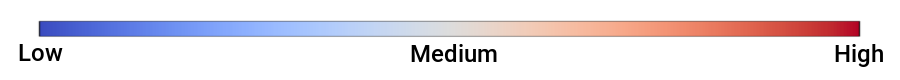}
    \end{subfigure}
    \caption{Measurement shift probability distribution: (a) normal distribution, (b) uniform distribution, (c) normal + uniform. Color corresponds to the measurement probability.}
    \label{fig:shift_probability}
\end{figure}

We assume a shift error to be a 2D Gaussian function with the mean centered in the real line position. In reality, the detection process usually produces a Gaussian noise of the detected line ends positions, so our model overestimates the noise. However, such an assumption significantly simplifies computation and allows for noise to be precomputed as it will be shown in the next section. Apart from the spatial detection noise, we account for false positives detection i.e. detection of lines when in reality there are none. To account for that we introduce additional uniform probability distribution of a random detection (see Fig.~\ref{fig:shift_probability}): 
\begin{equation}
    P^{k}_{\mathrm{shift}}( \mathbf{z} \mid \mathbf{x}, M) = \frac{1}{I} \sum_{i=1}^{I} \frac{1}{2\pi\sigma^2} e^{-\Big(\frac{h_x(p^{k}_{i})^2 + h_y(p^{k}_{i})^2}{2\sigma^2}\Big)} + \frac{1}{\alpha},
    \label{eq:shift_prob}
\end{equation}
where $I$ is a number of points approximating $k^{th}$ line, $h_x$ and $h_y$ are the $x^{th}$ and $y^{th}$ components of the vector between point $p^k_i$ in map reference frame and the closest point of the closest reference line, $\alpha$ is a parameter corresponding to the uniform distribution accounting for the false positive detection.  

To account for angular detection error the additional PDF is introduced. The likelihood of the detected line segment depends on the angle $\gamma$ between true line orientation and orientation of the detected segment. The angular noise is assumed to be Gaussian with zero mean: 
\begin{equation}
\begin{array}{l}
    P^{k}_{\mathrm{angle}}(\mathbf{z} \mid \mathbf{x}, M) = \frac{1}{I-1} \sum_{i=1}^{I-1} f\Big( |\gamma_i| \Big) = \\
    = \frac{1}{I-1} \sum_{i=1}^{I-1} \frac{1}{\sigma\sqrt{2\pi}}e^{-\frac{(|\gamma_i|)^2}{2\sigma^2}},
    \label{eq:angle_probability}
\end{array}
\end{equation}
where $I$ is a number of points approximating $k^{th}$ line.

In the case of the multi-camera system the likelihoods computed for separate cameras are multiplied in order to estimate total measurement likelihood:
\begin{equation}
    P( \mathbf{z} \mid \mathbf{x}, M) = \prod_{c} P( \mathbf{z_c} \mid \mathbf{x}, M),
    \label{eq:multi_camera_eq}
\end{equation}
where $c = 1...C, c \in \mathbb{Z}$ is a camera index.  

\section{MAP REPRESENTATION}

A digital map is required to compare the sensory data with the expected environment observation. The map is represented by a multichannel raster image where each channel contains different map-related information. The first channel stores the initial map of linear features used for the localization. It can be a map of road markings (see Fig.~\ref{fig:map_model}, top), schematic map of the building or map of any other linear features type.

The proposed measurement model allows the precomputing of the measurement model beforehand. The map stores precomputed components of the measurement model ($P_{\mathrm{shift}}$ and  $P_{\mathrm{angle}}$) in two separate map image channels.

The $P_{\mathrm{shift}}$ related channel is obtained from the initial map via Gaussian smoothing -- by convolving the original map image with 2D Gaussian kernel. The size of the kernel and standard deviation $\sigma$ depend on the detector shift noise, estimated experimentally and the map scale ($\frac{m}{pix}$). The uniform distribution $\frac{1}{\alpha}$ is simply added to each pixel after Gaussian smoothing (see Fig.~\ref{fig:map_model}, left). Therefore, computing the equation~\ref{eq:shift_prob} comes down to the summation of values stored in pixels, corresponding to the detected linear segments, transformed to the map reference frame.

The $P_{\mathrm{angle}}$ related channel is required to speed up computation of angular part of measurement model~(\ref{eq:angle_probability}). The angle $\gamma$ can be estimated if the shortest distances between ends of the detected segment and closest map segment are known. Such a distance can be precomputed. In image processing, there is a morphological operation known as distance transform~\cite{rosenfeld1968distance}. The result of the transform is a gray-scale image where each pixel stores a distance to the closest boundary (see Fig.~\ref{fig:map_model}, middle). Given the distances $d1$ and $d2$ of a line segment ends to the closest reference line, the absolute angular misalignment $|\gamma|$ is estimated as:
\begin{equation}
    |\gamma| = \arcsin \Big(\frac{|d1-d2|}{l_{d1,d2}}\Big), 
\end{equation}
where $l_{d1,d2}$ -- the length of the line segment.

Additional map channel represents occupancy grid map -- information about the areas where car presence is possible (see Fig.~\ref{fig:map_model}, right). This information allows discarding infeasible hypotheses, therefore, decreasing the computational complexity of the localization system.

Storing digital map in memory as a multichannel raster image allows accessing any location on the map in constant time $O(1)$ given that the area is limited. Such representation can also easily be extended with new prior information which might be added to an additional image channel.

\section{LOCALIZATION SYSTEM OVERVIEW}

\begin{figure}[thpb]
    \centering
    \includegraphics[width=0.4\linewidth]{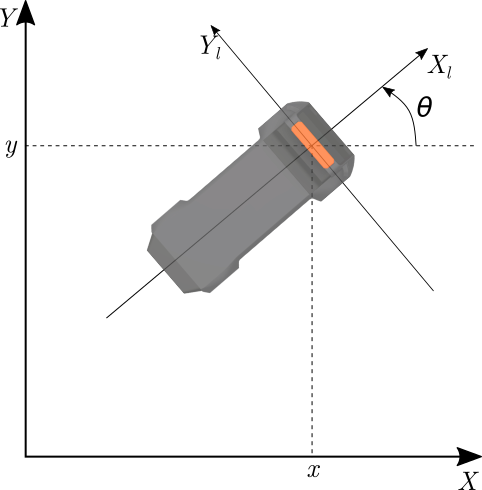}
    \caption{Car pose in map coordinate system.}
    \label{fig:map_coordinate_system}
\end{figure}

To evaluate the proposed measurement model the particle filter is used~\cite{dellaert1999monte}. It approximates the posterior distribution of the system state vector $ \mathbf{x_{t}} $ at the time moment $ t $ with a set of hypotheses (particles) with associated weights proportional to their likelihood estimations $S = \{ { \langle \mathbf{x^{n}_{t}}, w^{n}_{t}} \rangle \} $, where $ n = 1... N, n \in \mathbb{Z} $, $N$ -- total number of particles. Every particle is a hypothesis of the current vehicle pose $ \mathbf{x_{t}} = (x_{t}, y_{t}, \theta_{t}) $, where $ x_{t} $ and $ y_{t} $ are the 2D vehicle coordinates in the map reference frame at the time $ t $, and $ \theta_{t} $ is a corresponding yaw angle computed relative to the $ X $ axis of the map coordinate system (Fig.~\ref{fig:map_coordinate_system}). Unlike other localization methods, particle filter is able to handle non-linear models and multimodal probability distributions. 

We used a systematic resampling~\cite{kitagawa1996monte} at the resampling step of the particle filter. It outperforms other sampling approaches in terms of quality and computational complexity~\cite{hol2006resampling}.

\subsection{Motion Model}

Each filter iteration starts with the particles poses prediction based on the relative odometry measurements. Generally, proprioceptive sensors cannot provide reliable positioning on their own since their estimations are prone to the drift over time. Therefore only relative odometry measurements are applied to particles. To account for motion noise (e.g. imperfect actuators and odometry measurements) additive Gaussian noise with zero mean is applied both to the pose and orientation of each particle. Given the difference of two consecutive odometry measurements $(\Delta x_t, \Delta y_t,  \Delta \theta_t)$ actuation model can be expressed as follows: 
\begin{equation}
\left\{
\begin{array}{c}
    \theta^n_t = \theta^n_{t-1} + \Delta\theta_t^n + \Delta\theta_t^n \delta^n_t, \\
    x^n_t = x^n_{t-1} + d^n_t \cos (\theta^n_t), \\
    y^n_t = y^n_{t-1} + d^n_t \sin (\theta^n_t),
\end{array}
\right.
\end{equation}
where $ d^n_t = s_{t}  + s_{t} \eta^n_t $, $ s_{t} = \sqrt{(\Delta x_t)^2 + (\Delta y_t)^2} $, $ \delta^n_t = {\mathcal {N}}(0, \sigma_{\mathrm{angular}}^{2})$  and $ \eta^n_t = {\mathcal {N}}(0, \sigma_{\mathrm{linear}}^{2}) $ represent Gaussian noise with zero mean. Standard deviations of such distributions depend on the particular sensors set used for odometry. 

\subsection{Measurement Model}

To estimate likelihood (i.e. weight) of each particle the proposed measurement model is used. Detected relative to the vehicle pose linear features are approximated by polylines with constant segment length in the image reference frame. Polylines then transformed into the map reference frame for each particle individually, taking into account particle current pose on the map. Then the observation model (\ref{eq:meas_model}) is applied to estimate the likelihood of each particle.

In order to increase localization quality, prior information in the form of an occupancy grid map is used. It limits the area where particle presence is possible and therefore increases localization precision since only probable particles are considered. The likelihood function for the occupancy grid map is represented below:
\begin{equation}
{w_{om}}_t^n = \left\{ \begin{array}{ll}
    \text{1,} & \textrm{if pose is not occupied.}\\
    \text{0,} & \textrm{otherwise}.
 \end{array} \right.
\label{eq:w_om} 
\end{equation}
The particle likelihood computed according to (\ref{eq:meas_model}) is multiplied by the $w_{om}$. The estimated particles weights are then normalized:
\begin{equation}
   w_t^n = \frac{w_t^n}{\sum_{n=1}^{N} w_t^n} = \frac{P(\mathbf{z_t} \mid \mathbf{x_t^n}, M) {w_{om}}_t^n}{\sum_{n=1}^{N} P(\mathbf{z_t} \mid \mathbf{x_t^n}, M) {w_{om}}_t^n}.
\end{equation}
The resulting pose is computed as a weighted average of all particles.

\section{EXPERIMENTAL SETUP}

\begin{figure}[t]
    \centering
    \begin{subfigure}{0.25\textwidth}
        \includegraphics[width=\linewidth,keepaspectratio]{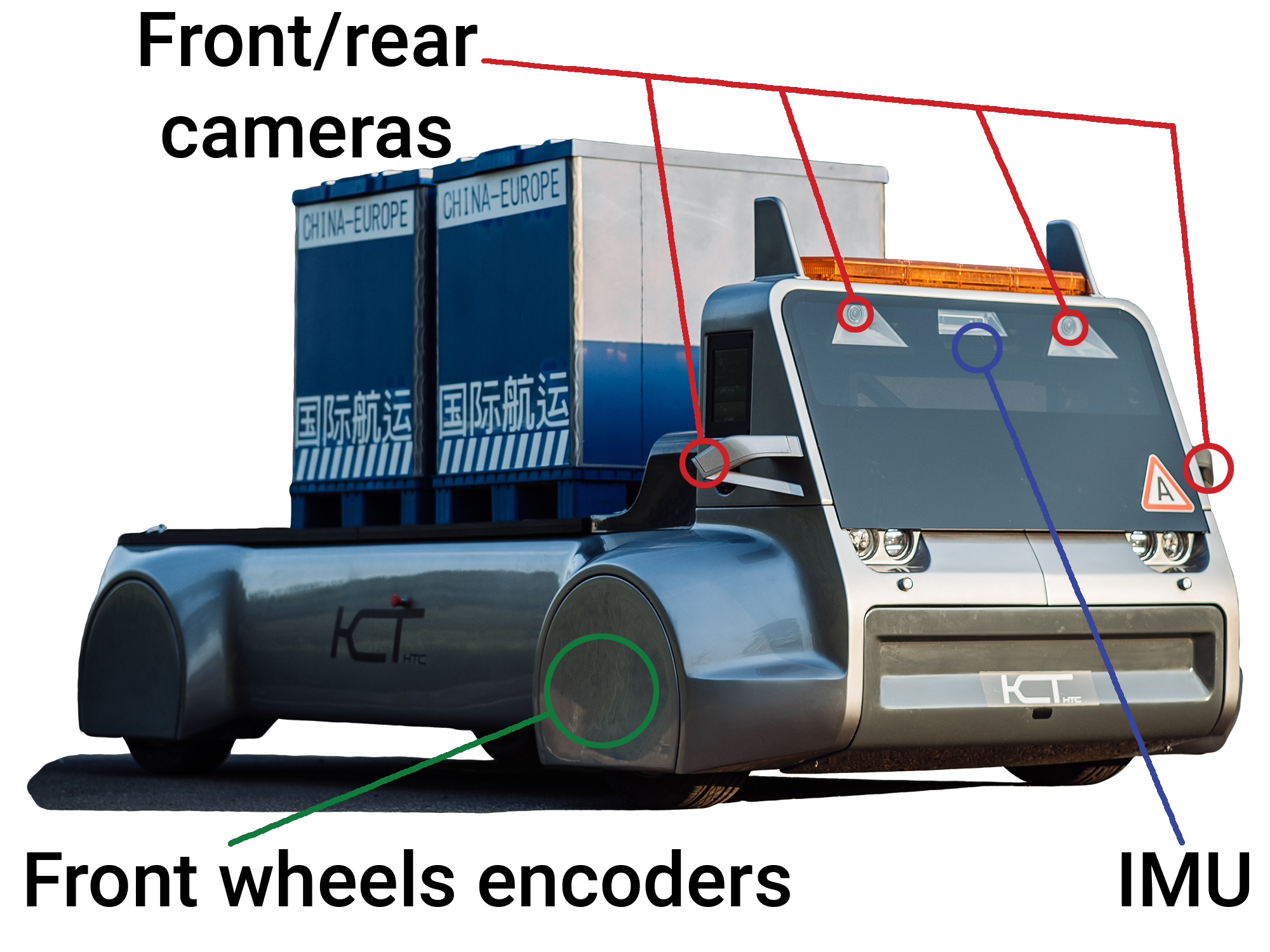}
        \caption{}
    \end{subfigure} 
    \hspace{10px}
    \begin{subfigure}{0.2\textwidth}
        \includegraphics[width=\linewidth,keepaspectratio]{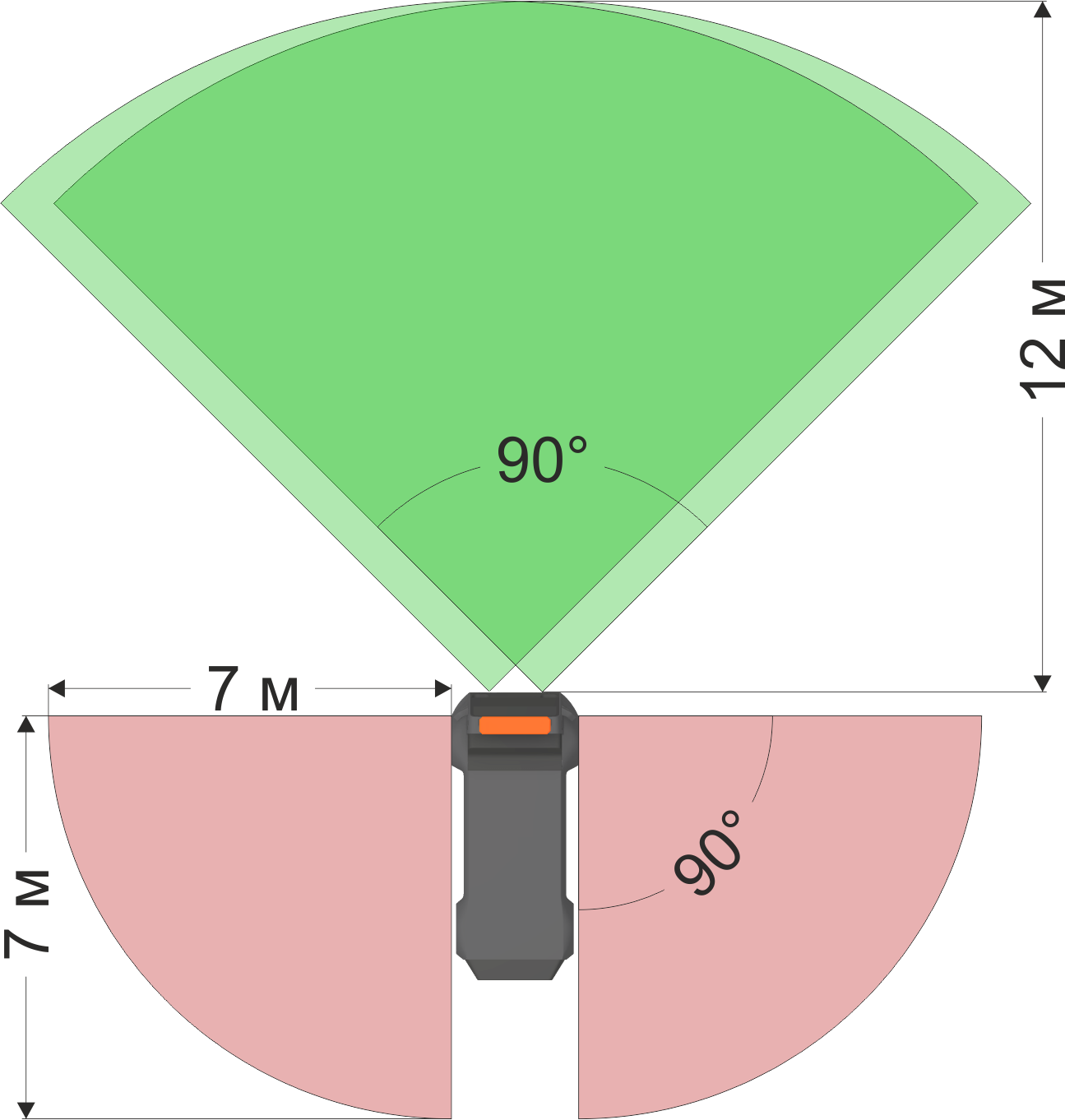}
        \caption{}
    \end{subfigure}
    \caption{The autonomous logistic vehicle used for measurement model evaluation:   (a) sensors placement, (b) cameras field of view.}
    \label{fig:car_platform}
\end{figure}   

\begin{figure}[thpb]
    \centering
    \begin{subfigure}{0.4\textwidth}
        \includegraphics[width=\linewidth,keepaspectratio]{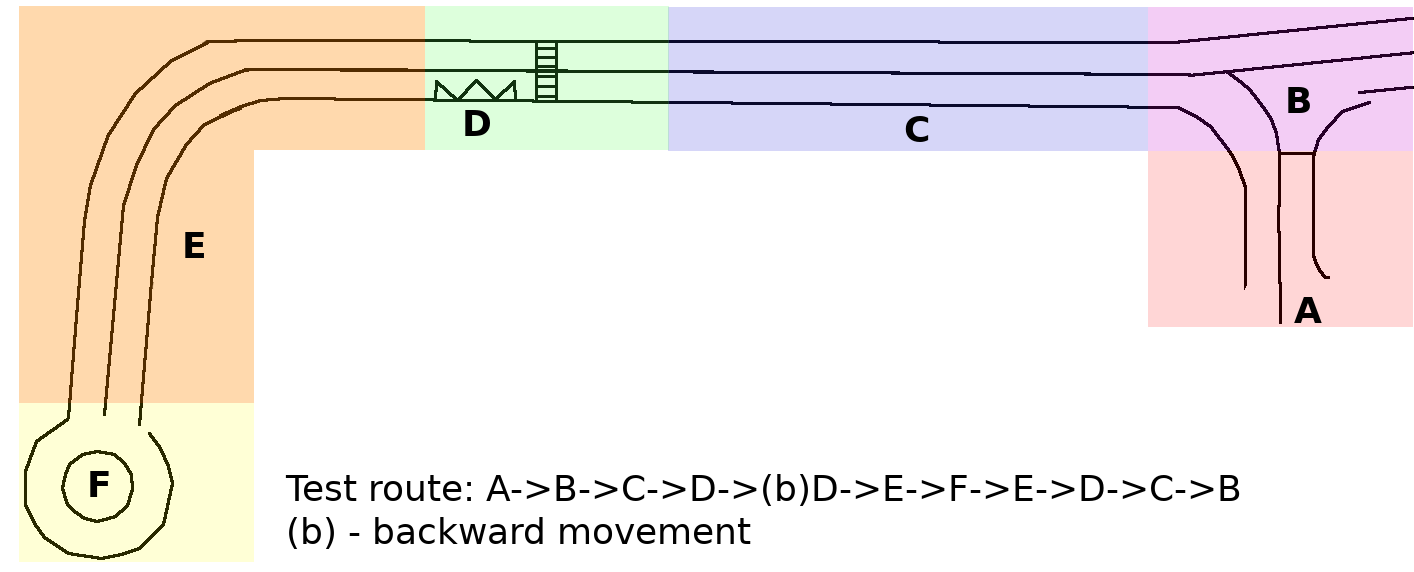}
        \caption{}
        \label{fig:prior_map}
    \end{subfigure} 
    \hspace{10px}
    \begin{subfigure}{0.5\textwidth}
        \includegraphics[width=\linewidth,keepaspectratio]{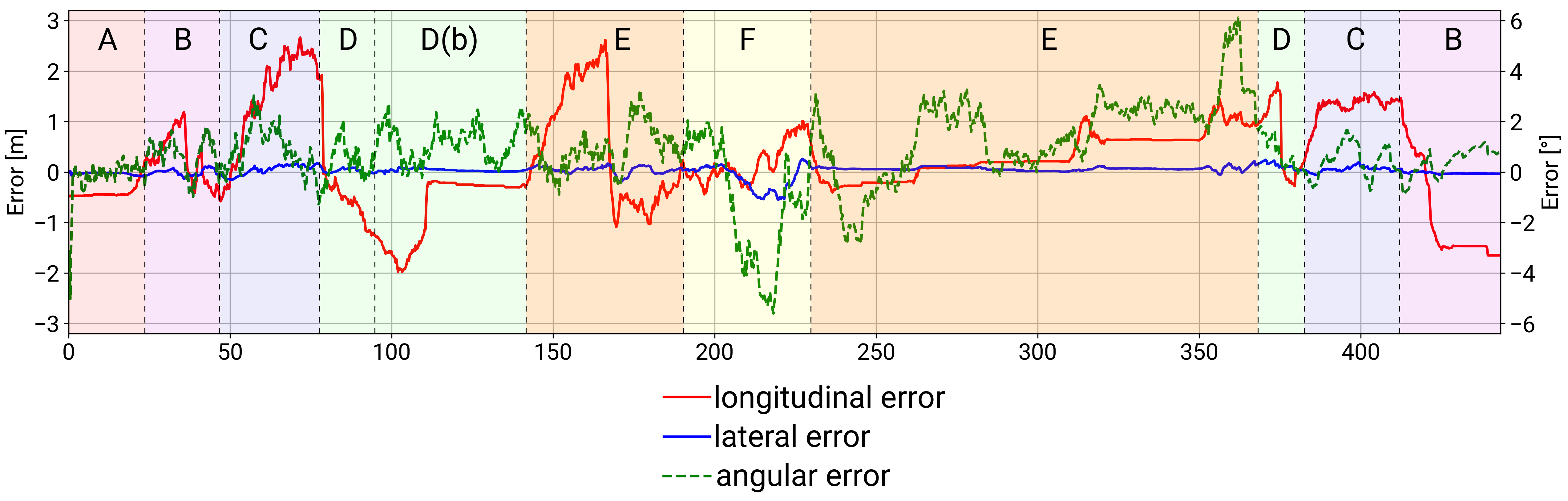}
        \caption{}
        \label{fig:results}
    \end{subfigure}
    \caption{Experimental results: (a) test route and road marking map, (b) errors dependence over time.}
    \label{fig:best_results}
\end{figure} 

This section details the results of the experimental evaluation of the proposed measurement model. It was evaluated in an outdoor environment on an autonomous logistic vehicle. The vehicle uses four cameras to detect road markings as shown in Fig.~\ref{fig:car_platform}. Fig.~\ref{fig:prior_map} demonstrates prior lane markings map of test site used to test localization performance as well as the test route which includes turns, straight and backward movement as well as a roundabout. The trajectory obtained with REACH RS+ RTK GNSS was used as a ground truth. To detect linear feature the Hough transform-based algorithm similar to the one described in~\cite{shipitko2019edge} was used. Both linear features detector and particle filter algorithm were implemented as C++ code and were running on the PC with Intel Core i7 CPU and 8 Gb RAM. No parallel processing was used. The number of particles used in all experiments was equal to $1000$.

The experiments were conducted to compare different observation models. The proposed model (shift+angular) was compared to the models accounting only for the shift error (shift) and only for angular error (angular). All numerical results were averaged over $10$ test runs.

\section{EXPERIMENTAL RESULTS}

The results are demonstrated in Table~\ref{tab:tab_result}. It can be seen that the localization with the proposed model is superior to the models only considering shift probability or angular misalignment probability. With the use of a combined model, the maximum absolute longitudinal error was decreased on $15\%$ in comparison with the model only considering measurement shift, maximum angular error on $19\%$ (Mean Absolute Error (MAE) on $15\%$). However, it needs to be noted that while lateral MAE was decreased on $12\%$, the maximum lateral error was increased from $0.49$ m to $0.55$ m (on $12\%$). 

Fig.~\ref{fig:results} depicts localization errors dependence on time for one of the test runs. It can be noted how lateral error growth on the straight segments of the route where there are no features for longitudinal correction (segments C, E). However, the error is immediately corrected once the distinctive features are observed -- road crossing between C and D (approximately $75$th seconds of the test run), turn in the middle of segment E. It also can be seen from Fig.~\ref{fig:results} that the angular error is positive most of the time. This bias can be due to the slightly wrong calibration of cameras or constant inclination of the car itself. 

\begin{figure}[t]
    \centering
    \includegraphics[width=0.9\linewidth,keepaspectratio]{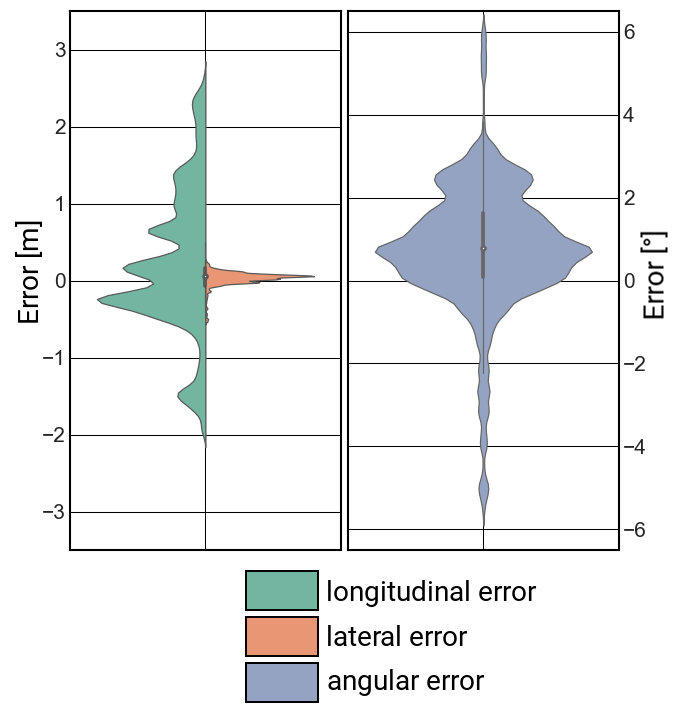}
    \caption{Localization errors distributions.}
    \label{fig:violin}
\end{figure}

Averaged over $10$ test runs distributions of longitudinal, lateral and angular errors for the proposed observation model are presented in Fig.~\ref{fig:violin}. It can be seen that the longitudinal error distribution is much wider. It leads to the conclusion that while lateral positioning can be efficiently performed based only on linear features the robust and precise longitudinal positioning requires additional sources of localization data. Such sources can be represented by other types of sensors or e.g. point features in case of visual localization. Fig.~\ref{fig:violin} also supports the hypothesis of bias in angular error due to some source of a static error.

\begin{table*}[thpb]
\footnotesize
\caption{Measured localization performance. Different rows correspond to different measurement models. Max stands for maximum error. MAE stands for Mean Absolute Error. } 
\label{tab:tab_result}
\begin{center}
\begin{tabular}{| c | c | c | c | c | c | c |}
 \hline
 \hline
  \multirow{2}{*}{\textbf{Model}} & \multicolumn{2}{|c|}{Longitudinal [m]} & \multicolumn{2}{|c|}{Lateral [m]} & \multicolumn{2}{|c|}{Angular [rad]} \\ \cline{2-7}
  & Max & MAE & Max & MAE & Max & MAE \\ 
 \hline
 shift & $3.07$ & $\mathbf{0.72 \pm 0.92}$ & $\mathbf{0.49}$ & $0.08 \pm 0.11$ & $7.59$ & $1.53 \pm 1.65$ \\ 
 angular & $2.77$ & $0.86 \pm 0.94$ & $2.15$ & $0.69 \pm 0.64$ & $8.99$ & $1.97 \pm 1.9$ \\  
 shift + angular & $\mathbf{2.66}$ & $\mathbf{0.72 \pm 0.92}$ & $0.55$ & $\mathbf{0.07 \pm 0.1}$ & $\mathbf{6.11}$ & $\mathbf{1.29 \pm 1.51}$ \\    
 \hline
 \hline
\end{tabular}
\end{center}
\end{table*}

\begin{figure}[thpb]
    \centering
    \includegraphics[width=\linewidth,keepaspectratio]{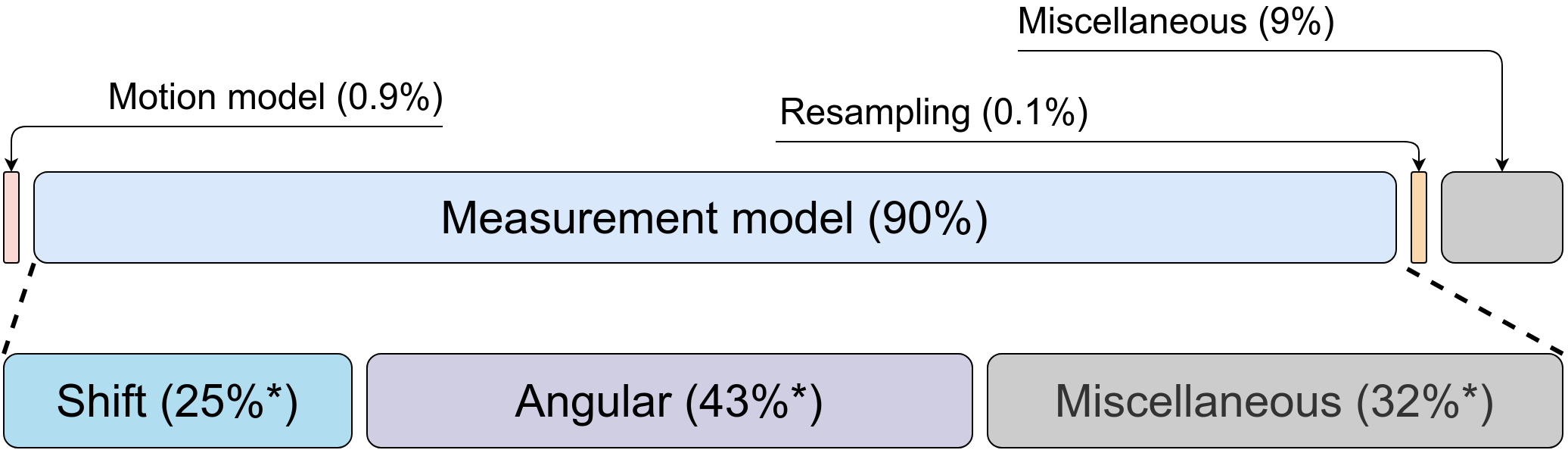}
    \caption{Time profiling of the single particle filter iteration. Note, the percentage marked with \textit{asterisk} symbol is percentage form the observation model, not the whole particle filter iteration. The whole iteration takes on average 0.012 seconds.}
    \label{fig:time_profiling}
\end{figure}

To analyze computational load introduced with a model modification that considers angular misalignment we performed an algorithm computation time profiling. The results are demonstrated in Fig.~\ref{fig:time_profiling}. It can be seen that the model component corresponding to angular error takes twice the time of shift component to be computed. Nevertheless, the whole particle filter step including measurement model computation is very fast (can run at $\approx83$ Hz) and can be further accelerated by parallelized computations.  

\section{CONCLUSION}

In this work, the probabilistic observation model for visually detected linear features is proposed. The proposed model can be used in localization algorithms such as particle filter and Kalman filter. The proposed model accounts for three measurement error sources: detection shift error, orientation error, and false-positive detection. The model overcomes known drawback of classical models -- their bias towards incorrect orientation estimation. The structure of the model allows to directly incorporate a precomputed measurement model into the map which speeds up the localization algorithm.

The proposed model was evaluated on a real autonomous vehicle. Experimental results have demonstrated the superiority of the proposed model over the traditional model considering only a shift between expected and obtained measurements. We also evaluated the computational load required by the proposed extended model. We showed that although it doubles computational requirements the proposed model still can be efficiently computed even on the low-end processing unit.

The future work may include the expansion of the proposed model to incorporate external error sources (e.g. increasing detection uncertainty with the increasing detection distance or uneven road surface) and internal error sources (e.g. unpredicted vehicle motions or error model of particular underlying linear features detection algorithm). The future work also includes the application of the proposed model to other real-life localization scenarios such as indoor localization using linear features.



\bibliographystyle{IEEEtran}
\bibliography{mybib}

\newpage

\copyright{} 2020 IEEE.  Personal use of this material is permitted.  Permission from IEEE must be obtained for all other uses, in any current or future media, including reprinting/republishing this material for advertising or promotional purposes, creating new collective works, for resale or redistribution to servers or lists, or reuse of any copyrighted component of this work in other works.

\end{document}